%% file: main.tex
\documentclass[10pt,a4paper]{article}

\usepackage[margin=2.5cm]{geometry}
\usepackage{graphicx}
\usepackage{booktabs}
\usepackage{array}
\usepackage{longtable}
\usepackage{multirow}
\usepackage{amsmath}
\usepackage{amssymb}
\usepackage{hyperref}
\usepackage{xcolor}
\usepackage[super,sort&compress]{natbib}
\usepackage{authblk}
\usepackage{setspace}
\usepackage{caption}
\usepackage{subcaption}
\usepackage{lineno}
\usepackage{microtype}

\hypersetup{
    colorlinks=true,
    linkcolor=blue,
    citecolor=blue,
    urlcolor=blue
}

\title{\textbf{A Vision Transformer for ECG-Based Detection of Left Ventricular Systolic Dysfunction Across Multiple Clinical Sites}}

\author[1,*]{Burcu Ozek}
\author[1]{Aruna Mohan}
\author[4]{David Vorchheimer}
\author[3]{Daniel Weiss}
\author[2]{Eyal Kedar}
\author[1]{Tamar Sobol}
\author[1,*]{Or Zilbershot}
\author[5,*]{Fatemeh Afghah}

\affil[1]{Walkky LLC, Boston, MA, USA}
\affil[2]{St. Lawrence Health, Canton, NY, USA}
\affil[3]{Weiss Consulting Groups, Boca Raton, FL, USA}
\affil[4]{Northwell Cardiovascular Institute, Lenox Hill Hospital, New York, NY, USA}
\affil[5]{Clemson University, Clemson, SC, USA}
\affil[*]{Corresponding authors: Burcu Ozek (burcu@walkky.com); Or Zilbershot (or@walkky.com); Fatemeh Afghah (fatemeh@walkky.com)}

\date{}

\begin{document}

\maketitle

% ── Abstract ──────────────────────────────────────────────────────────────────
\begin{abstract}
\input{sections/abstract}
\end{abstract}


\newpage

% ── Main Sections ─────────────────────────────────────────────────────────────
\input{sections/introduction}
\input{sections/results}

\input{sections/discussion}
\input{sections/methods}

% ── End Matter ────────────────────────────────────────────────────────────────

\section*{Data Availability}
The data analyzed in this study are not publicly available owing to data licensing agreements and patient privacy protections. Data were accessed through the Dandelion Health platform and may be made available to qualified researchers upon reasonable request to Dandelion Health, subject to their data use agreements.

\section*{Code Availability}
The underlying code used to train and evaluate the models in this study is not publicly available for proprietary reasons.

\section*{Author Contributions}
B.O.: Conceptualization, Methodology, Software, Data curation, Formal analysis, Validation, Visualization, Investigation, Writing (original draft, review, and editing).
A.M.: Conceptualization, Methodology, Software, Investigation, Validation, Formal analysis, Writing (review and editing).
D.V., D.W., E.K., and T.S.: Writing (review and editing).
O.Z.: Conceptualization, Resources, Project administration, Writing (review and editing).
F.A.: Conceptualization, Supervision, Methodology, Writing (review and editing).

\section*{Competing Interests}
The authors declare no competing interests.

\section*{Acknowledgements}
This study received no funding.

% ── References ────────────────────────────────────────────────────────────────
\bibliographystyle{naturemag}
\bibliography{references}

% ── Appendix ──────────────────────────────────────────────────────────────────
\newpage
\appendix
\input{sections/appendix}

\end{document}

%% file: sections/abstract.tex
% ══════════════════════════════════════════════════════════════════════════════
% ABSTRACT — unstructured, ~183 words (npj Digital Medicine Article style; limit 150)
% ══════════════════════════════════════════════════════════════════════════════

Reduced left ventricular ejection fraction (LVEF) is frequently asymptomatic and
often detected only after advanced heart failure develops. Electrocardiograms are
recorded routinely yet underused for this condition, because reduced LVEF has no
single diagnostic waveform. We trained an ensemble of vision transformers from
scratch to detect reduced LVEF ($\leq$40\%) from 12-lead ECGs, analyzing each
heartbeat individually, using 10,142 patients across seven sites in three US health
systems. In a held-out external cohort of 4,092 patients from three geographically
independent US clinical sites at a real-world reduced-LVEF prevalence of 8.72\%, the model
achieved an AUROC of 0.88 (95\% CI 0.86--0.89), sensitivity 81.2\%, specificity
81.0\%, and negative predictive value 97.8\%. Sensitivity remained high across sex,
race, ethnicity, and comorbidity subgroups, while specificity was lower in older
patients and those with atrial fibrillation or cardiomyopathy. Beat-level attention
maps provided interpretability into the model's predictions, showing consistent focus
on the QRS complex rather than the P wave. These findings support the potential of
routine ECGs as a scalable first-pass triage step to identify patients who should
undergo echocardiography for reduced ejection fraction across diverse patient
populations.

%% file: sections/introduction.tex
% ══════════════════════════════════════════════════════════════════════════════
% INTRODUCTION
% ══════════════════════════════════════════════════════════════════════════════

\section*{Introduction}

Left ventricular systolic dysfunction (LVSD), defined as a left ventricular
ejection fraction at or below 40\%~\cite{heidenreich2022,mcdonagh2023}, has
an estimated prevalence of 3--6\% in the general population. It frequently causes
no symptoms and often goes undetected until symptomatic heart failure is
established~\cite{wang2003}. Because early LVSD may be asymptomatic or present with nonspecific symptoms~\cite{yeboah2012}, many patients are diagnosed only after progression to overt heart failure, when opportunities for earlier initiation of guideline-directed therapy may have been missed~\cite{heidenreich2022}. Transthoracic echocardiography remains the clinical reference standard for measuring LVEF; however, in most settings its use as a broad screening tool remains constrained by cost, limited availability, the need for trained personnel, dependence on image quality and expert interpretation, and workflow burden~\cite{bjerkn2023,attia2019}. These barriers are especially relevant in primary care, emergency, rural, and resource-limited settings, where routine echocardiographic screening of large at-risk populations is not practical. By contrast, the 12-lead electrocardiogram is inexpensive, rapid, widely available, and routinely obtained across nearly all clinical settings. However, ECG-based detection of reduced LVEF is not a traditional ECG interpretation task. Unlike arrhythmias, conduction blocks, or acute ischemic changes, LVSD does not produce a single diagnostic waveform pattern that can be reliably identified by predefined rules or visual inspection. Its electrical manifestations are often subtle, distributed across leads, and influenced by comorbidities such as atrial fibrillation, myocardial infarction, cardiomyopathy, and hypertension. These signals may reflect complex interactions among ventricular remodeling, conduction delay, repolarization abnormalities, and myocardial disease. As a result, routine clinical ECG interpretation has historically not been used to screen for reduced LVEF, despite the possibility that latent information about ventricular function is embedded in the waveform~\cite{ose2024,rhee2026,kligfield2007}.

Artificial intelligence offers a way to extract these latent, high-dimensional ECG signatures and transform a routinely acquired test into a scalable approach to flag patients for echocardiographic evaluation of reduced LVEF. Convolutional neural networks (CNNs) applied to the 12-lead ECG have
demonstrated that routine ECG recordings carry clinically meaningful
information for detecting reduced ejection fraction. The approach was introduced by Attia et al., who trained a CNN at a
single tertiary center and reported strong discriminative performance on an
internal test set~\cite{attia2019}. More recently, Carter et al.\ conducted a multi-site external
validation of an FDA-cleared version of this algorithm across four
geographically diverse US institutions and confirmed its performance at real-world
disease prevalence~\cite{carter2026}. CNN-based models have since been applied
in emergency department triage~\cite{adedinsewo2020}, general population
screening~\cite{rhee2026, killalea2026, lin2026}, and pediatric and congenital heart disease
populations~\cite{mayourian2025}. CNNs rely on local convolutional filters with limited receptive fields. This
makes them effective at capturing local waveform morphology but less suited to
modeling long-range temporal relationships across the heartbeat. Transformer-based
architectures address this through self-attention, which directly relates distant
regions of the signal~\cite{sawano2024,moody2025}. Applied to individual
heartbeats, this approach combines local detail within each patch with long-range
dependencies across the cardiac cycle, while the same attention reveals which ECG
regions drive the prediction~\cite{mohan2024}. The largest ECG models, however, rely on
extensive pretraining; HeartBEiT used 8.5 million ECGs from 2.1 million
patients~\cite{heartbeit2023}, ECGFounder over 10 million
recordings~\cite{ecgfounder2025}, and RhythmBERT 800,000~\cite{wang2026rhythmbert},
data scales beyond the reach of most institutions. In addition, reported predictive values in this literature are often derived
from cohorts enriched for disease, at prevalence higher than in routine
practice~\cite{heartbeit2023}.

We developed a vision transformer (ViT) ensemble, trained from scratch, to detect reduced LVEF ($\leq$40\%) from individual heartbeats. The model was trained on 10,142 patients who underwent paired ECG and echocardiography across multiple care settings at three US health systems. It was validated internally on 3,295 patients and externally on 4,092 patients from three fully held-out sites, at a real-world prevalence of 8.72\%. Under identical training data and splits, it outperformed a reimplementation of the established CNN architecture for ECG-based LVEF detection. Beat-level attention maps provided interpretability into the ECG features driving its predictions. We further assessed robustness through subgroup analyses, clinical utility through decision curve analysis, calibration of predicted probabilities, and the contribution of each training component through a systematic ablation. These results show that a ViT trained on a clinically accessible dataset can serve as an effective first-pass triage tool to guide echocardiography referral for reduced LVEF.

%% file: sections/results.tex
% ══════════════════════════════════════════════════════════════════════════════
% RESULTS
% ═══════════════════════════.       ═══════════════════════════════════════════════════

\section*{Results}
% ── 5.1 Study Cohort ──────────────────────────────────────────────────────────

\subsection*{Study Cohort}

A total of 17,529 patients were partitioned into training, internal validation, and external validation sets.
The dataset spanned 10 clinical sites across three US health systems; 7 sites contributed patients to both the training and internal validation sets, and 3 sites, one from each health system, were reserved exclusively for external validation with no involvement in model development.
The training set comprised 10,142 patients.
The internal validation set comprised 3,295 patients (8.71\% LVEF $\leq 40\%$)
from 7 clinical sites, and the external validation set comprised 4,092 patients
(8.72\% LVEF $\leq 40\%$) from 3 geographically independent clinical sites.
Patient characteristics for both validation sets are summarized in
Table~\ref{tab:demographics}.

In the external validation set, the mean age was 65.2 years (SD 16.6)
and 50.1\% were female; the racial and ethnic composition and encounter-setting
distribution are detailed in Table~\ref{tab:demographics}.
Each of the three external validation sites was a distinct acquisition location
not used during model development: one from Sanford Health (n\,=\,1,500), one
from Sharp HealthCare (n\,=\,1,342), and one from Texas Health Resources
(n\,=\,1,250).

\begin{table}[htbp]
\centering
\caption{\textbf{Patient characteristics of the internal and external validation sets.}}
\label{tab:demographics}
\begin{tabular}{lcc}
\toprule
\textbf{Characteristic} & \textbf{Internal Validation} & \textbf{External Validation} \\
                        & \textbf{(n\,=\,3,295)}       & \textbf{(n\,=\,4,092)}       \\
\midrule
LVEF $\leq 40\%$, \% (n)     & 8.71\% (287)  & 8.72\% (357)  \\
Age, mean (SD), years        & 66.5 (15.8)   & 65.2 (16.6)   \\
Female, \%                   & 51.1\%        & 50.1\%        \\
\addlinespace
\textit{Race/ethnicity, \%} & & \\
\quad White, non-Hispanic    & 51.1\%        & 62.3\%        \\
\quad Hispanic               & 18.9\%        & 20.0\%        \\
\quad Black/African American & 11.8\%        & 10.8\%        \\
\quad Asian                  & 7.8\%         & 3.1\%         \\
\quad Other/Multiple         & 10.5\%        & 3.8\%         \\
\addlinespace
\textit{Encounter setting, \%} & & \\
\quad Inpatient              & 45.1\%        & 45.5\%        \\
\quad Outpatient             & 36.6\%        & 42.6\%        \\
\quad Observation/Other      & 18.3\%        & 11.9\%        \\
\bottomrule
\end{tabular}
\end{table}

% ── 5.2 Model Performance ─────────────────────────────────────────────────────

\subsection*{Model Performance}

\begin{table}[htbp]
\centering
\caption{\textbf{ViT ensemble performance on internal and external validation sets.}
All metrics computed at threshold = 0.45. 95\% CIs from BCa bootstrap
(10,000 resamples).}
\label{tab:main_results}
\renewcommand{\arraystretch}{1.15}
\begin{tabular}{lcc}
\toprule
\textbf{Metric} &
\textbf{Internal Validation} & \textbf{External Validation} \\
& \textbf{(n\,=\,3,295)} & \textbf{(n\,=\,4,092)} \\
\midrule
AUROC        & 0.880 (0.859--0.897) & 0.878 (0.858--0.894) \\
Sensitivity  & 82.9\% (78.2--87.0) & 81.2\% (76.9--85.0) \\
Specificity  & 80.4\% (78.9--81.8) & 81.0\% (79.8--82.3) \\
PPV          & 28.7\% (25.7--31.8) & 29.1\% (26.3--32.0) \\
NPV          & 98.0\% (97.4--98.5) & 97.8\% (97.3--98.3) \\
F1 Score     & 42.7\% (39.0--46.2) & 42.8\% (39.5--46.1) \\
Accuracy     & 80.6\% (79.2--81.9) & 81.1\% (79.9--82.3) \\
\bottomrule
\end{tabular}
\end{table}

\subsubsection*{Internal Validation}

On the internal validation set, the ViT ensemble achieved an AUROC of 0.880
(95\% CI: 0.859--0.897), a sensitivity of 82.9\% (95\% CI: 78.2--87.0\%), and a
specificity of 80.4\% (95\% CI: 78.9--81.8\%) at the pre-specified threshold of
0.45 (Table~\ref{tab:main_results}).

\subsubsection*{External Validation}

On the held-out external validation set, performance was consistent with the
internal results (Table~\ref{tab:main_results}).
The ViT ensemble achieved an AUROC of 0.878 (95\% CI: 0.858--0.894), a
sensitivity of 81.2\% (95\% CI: 76.9--85.0\%), and a specificity of 81.0\%
(95\% CI: 79.8--82.3\%).
The NPV was 97.8\% (95\% CI: 97.3--98.3\%), indicating that among patients
identified as low-risk by the model, about 2 in 100 had true LVEF
$\leq 40\%$.
The PPV was 29.1\% (95\% CI: 26.3--32.0\%).

% ── 5.x Calibration ───────────────────────────────────────────────────────────

\subsection*{Calibration}

\begin{figure}[htbp]
    \centering
    \includegraphics[width=0.6\textwidth]{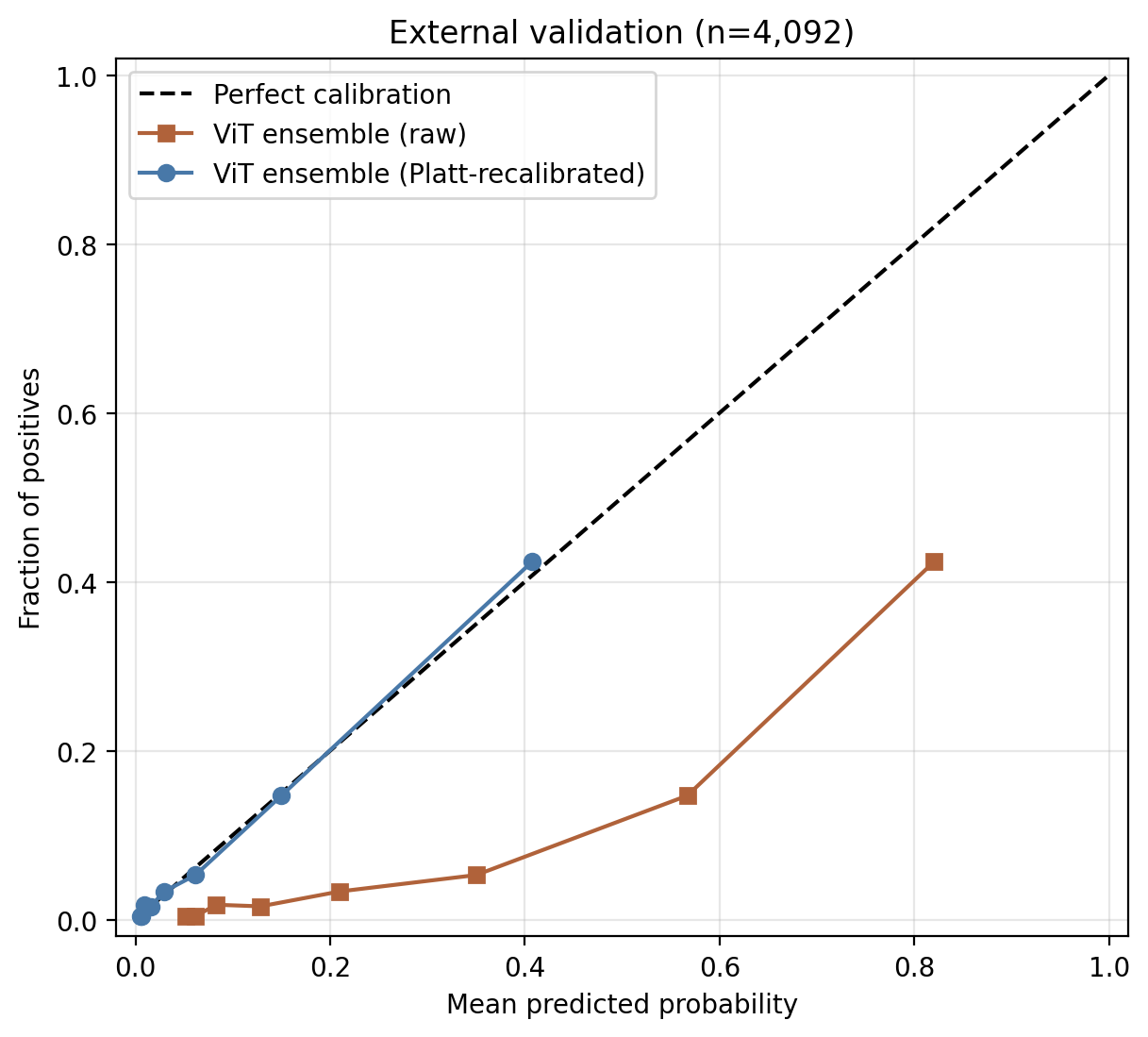}
    \caption{\textbf{Calibration of the ViT ensemble on the external validation
    set (n\,=\,4,092).}
    The raw ensemble (orange) overestimated risk, falling below
    the diagonal. Post-hoc Platt scaling (blue), fit on the internal validation
    set and applied to the external set, restored calibration to the
    diagonal without affecting discrimination (AUROC) or the binary operating
    point.}
    \label{fig:calib}
\end{figure}

On the external validation set, the raw ensemble systematically overestimated
absolute risk. The mean predicted probability was 28.4\% against an observed
prevalence of 8.72\%, with a Brier score of 0.121 (Figure~\ref{fig:calib}).
The calibration slope was 1.06 (ideal 1.0), indicating that the spread of
predicted probabilities was appropriate, whereas the calibration-in-the-large
was $-2.01$ (ideal 0.0), reflecting a uniform upward shift in predicted risk
rather than a failure of discrimination. This pattern is the expected
consequence of the minority oversampling and class weighting used during
training.

Post-hoc Platt scaling, fit on the internal validation set and applied to the
external set, corrected this shift. After recalibration, the mean predicted
probability fell to 8.6\%, the Brier score improved to 0.061, the calibration
slope was 0.99, and the calibration-in-the-large was 0.02
(Figure~\ref{fig:calib}). Discrimination was unchanged (AUROC 0.878), and
because the 0.45 operating threshold maps to an equivalent threshold on the
calibrated scale, the same patients were flagged, leaving sensitivity (81.2\%)
and specificity (81.0\%) identical.

% ── 5.5 Decision Curve Analysis ───────────────────────────────────────────────

% ── 5.7 Subgroup Analysis ─────────────────────────────────────────────────────

\subsection*{Subgroup Analysis}

Subgroup performance on the external validation set is shown in
Table~\ref{tab:subgroup}.
Performance was consistent across sex, with sensitivity of 81.4\% in males
and 81.0\% in females (AUROC 0.873 and 0.879, respectively).
Specificity was highest in patients under 60 (89.3\%) and lowest in those over
75 (72.3\%), while sensitivity was highest in the oldest group (86.6\%).
Among racial and ethnic groups with sufficient sample sizes, AUROC ranged from
0.849 in White, non-Hispanic patients to 0.918 in Black or African American
patients. Hispanic patients similarly showed strong performance, with a sensitivity
of 83.1\%, specificity of 86.9\%, and AUROC of 0.917.
Asian, American Indian or Alaska Native, Native Hawaiian or Other Pacific Islander,
and Other or Multiple Race patients showed 100\% sensitivity, though these estimates
should be interpreted with caution given the small number of positive cases in each group.

AUROC was 0.917 at Sharp HealthCare, 0.873 at Sanford Health, and 0.836
at Texas Health Resources.
Performance was similar between inpatient and outpatient settings
(AUROC 0.866 vs.\ 0.889).

Sensitivity ranged from 79.6\% to 85.9\% across all comorbidity subgroups,
indicating that the model reliably detects reduced ejection fraction regardless
of comorbidity burden.
Specificity was reduced in patients with cardiomyopathy
(57.0\%), heart failure (66.0\%), myocardial infarction (69.0\%), and atrial
fibrillation (70.2\%). Performance in patients with hypertension, diabetes, and
obesity remained strong and consistent with the overall cohort.

\renewcommand{\arraystretch}{1.1}
\setlength{\tabcolsep}{4pt}
\small
\begin{longtable}{p{2.2cm}p{4.4cm}rrrrr}
\caption{\textbf{Subgroup performance on the external validation set.}
All metrics computed at threshold = 0.45.
Prev = LVEF $\leq 40\%$ prevalence within subgroup.
Sens = sensitivity; Spec = specificity; AUC = AUROC.
$*$ Fewer than 10 positive cases; estimates should be interpreted with caution.}
\label{tab:subgroup} \\
\toprule
\textbf{Category} & \textbf{Subgroup} & \textbf{n} & \textbf{Prev (\%)} & \textbf{Sens (\%)} & \textbf{Spec (\%)} & \textbf{AUC} \\
\midrule
\endfirsthead
\multicolumn{7}{l}{\textit{(continued)}} \\
\toprule
\textbf{Category} & \textbf{Subgroup} & \textbf{n} & \textbf{Prev (\%)} & \textbf{Sens (\%)} & \textbf{Spec (\%)} & \textbf{AUC} \\
\midrule
\endhead
\midrule
\multicolumn{7}{r}{\textit{Continued on next page}} \\
\endfoot
\bottomrule
\endlastfoot
Sex              & Male                                      & 2,042 & 11.3 &  81.4 & 79.1 & 0.873 \\
                 & Female                                    & 2,050 &  6.1 &  81.0 & 82.8 & 0.879 \\
\midrule
Age              & $<$60 years                               & 1,349 &  8.2 &  80.0 & 89.3 & 0.918 \\
                 & 60--75 years                              & 1,465 &  8.7 &  77.3 & 81.0 & 0.863 \\
                 & $>$75 years                               & 1,278 &  9.3 &  86.6 & 72.3 & 0.850 \\
\midrule
Race/Ethnicity   & White, non-Hispanic                       & 2,548 &  8.2 &  77.5 & 79.6 & 0.849 \\
                 & Hispanic                                  &   820 &  8.7 &  83.1 & 86.9 & 0.917 \\
                 & Black or African American, non-Hispanic   &   442 & 12.9 &  87.7 & 79.7 & 0.918 \\
                 & Asian, non-Hispanic$*$                       &   125 &  5.6 & 100.0 & 82.2 & 0.939 \\
                 & American Indian or Alaska Native$*$          &    46 & 10.9 & 100.0 & 75.6 & 0.859 \\
                 & Native Hawaiian or Other Pacific Islander$*$ &    11 &  9.1 & 100.0 & 80.0 & 0.900 \\
                 & Other or Multiple Race, non-Hispanic$*$      &    70 &  7.1 & 100.0 & 80.0 & 0.972 \\
                 & Unknown/Declined$*$                          &    30 &  6.7 &  50.0 & 64.3 & 0.786 \\
\midrule
Health System    & Sanford Health                            & 1,500 &  8.0 &  76.7 & 83.0 & 0.873 \\
                 & Sharp HealthCare                          & 1,342 &  9.3 &  88.8 & 81.8 & 0.917 \\
                 & Texas Health Resources                    & 1,250 &  9.0 &  77.7 & 77.9 & 0.836 \\
\midrule
Encounter        & Inpatient                                 & 1,864 & 11.1 &  81.6 & 78.0 & 0.866 \\
Setting          & Outpatient                                & 1,742 &  6.5 &  81.4 & 83.1 & 0.889 \\
                 & Observation$*$                              &   384 &  2.3 &  66.7 & 88.3 & 0.852 \\
                 & Other in person$*$                          &    29 & 13.8 &  75.0 & 56.0 & 0.620 \\
                 & Other/Unspecified$*$                        &    11 & 18.2 &  50.0 & 33.3 & 0.278 \\
\midrule
Comorbidities    & Heart failure                             & 1,473 & 22.1 &  85.5 & 66.0 & 0.817 \\
                 & Atrial fibrillation                       & 1,232 & 13.1 &  83.2 & 70.2 & 0.830 \\
                 & Cardiomyopathy                            &   393 & 43.3 &  85.9 & 57.0 & 0.788 \\
                 & Myocardial infarction                     &   781 & 19.5 &  79.6 & 69.0 & 0.799 \\
                 & Coronary artery disease                   & 1,629 & 13.8 &  82.7 & 74.7 & 0.843 \\
                 & Hypertension                              & 2,955 &  8.8 &  79.9 & 79.4 & 0.861 \\
                 & Diabetes                                  & 1,534 & 10.0 &  85.7 & 76.7 & 0.874 \\
                 & Obesity                                   & 1,438 &  7.7 &  82.0 & 82.3 & 0.879 \\
\end{longtable}

% ── 5.6 Attention Maps ────────────────────────────────────────────────────────

\subsection*{Attention Maps}

\begin{figure}[htbp]
    \centering
    \includegraphics[width=\textwidth]{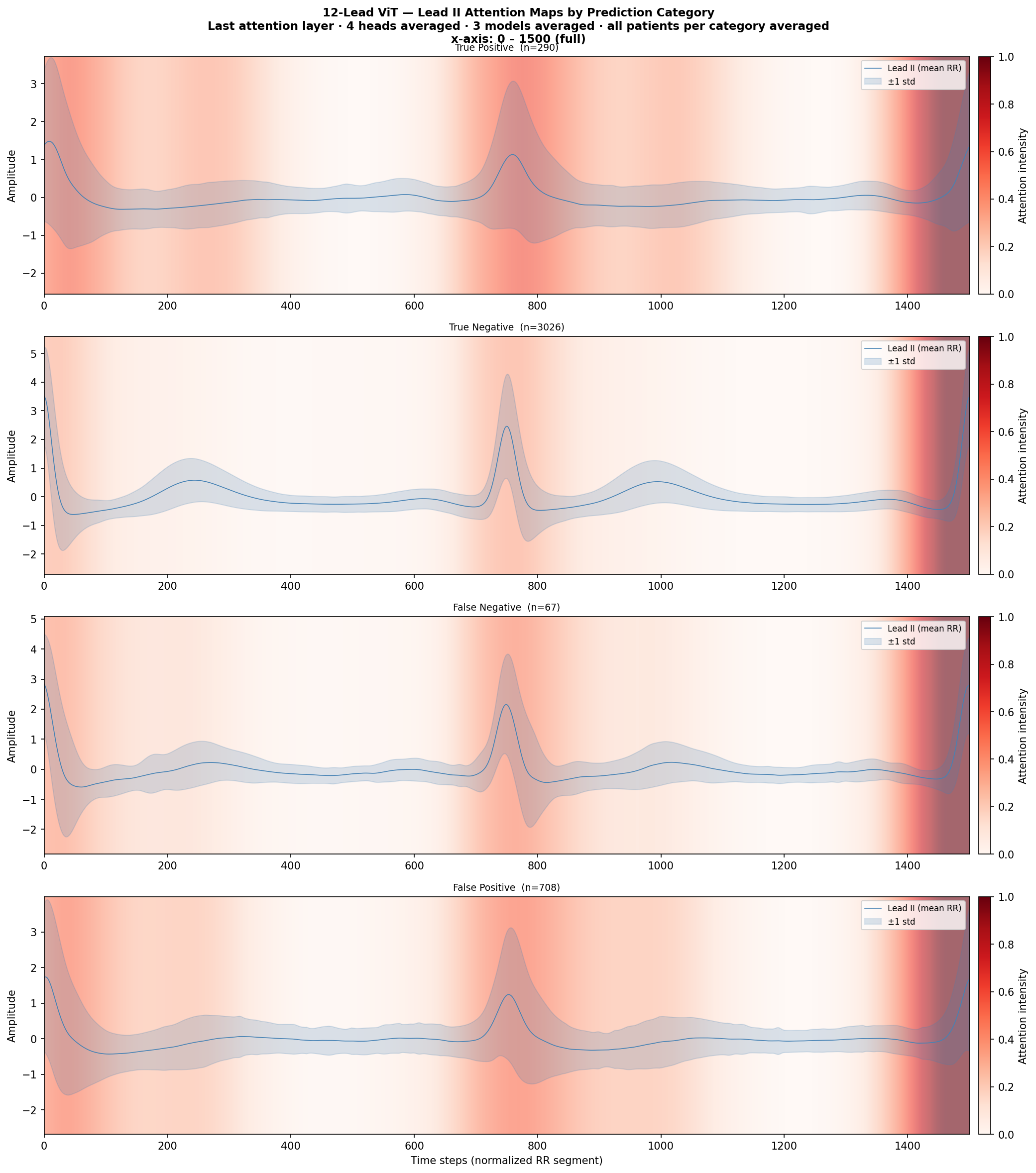}
    \caption{\textbf{Attention maps of the 12-lead ViT by prediction category.}
    These maps show which parts of each heartbeat the model learned to focus on, rather than relying on predefined rules.
    Attention weights were extracted from the final transformer encoder layer,
    averaged across 4 attention heads, 3 ensemble models, and all patients per
    category. The Lead~II signal (mean $\pm$ 1 SD) is overlaid on the attention
    heatmap (red intensity). Attention weights are derived from the full 12-lead
    input; the Lead~II signal is shown for display purposes only.
    TP = true positive; TN = true negative; FN = false negative; FP = false positive.}
    \label{fig:attention}
\end{figure}

Attention maps extracted from the final transformer encoder layer are shown in
Figure~\ref{fig:attention}, stratified by prediction category (true positive,
true negative, false negative, false positive).
In true-positive cases, attention was concentrated over the QRS complexes, which
fall at the edges and center of the beat window, with weaker, secondary attention
extending into the adjacent ST-T segment. False-negative cases showed attention concentrated on the QRS complex, with
little attention over the adjacent ST-T segment and T wave. False-positive cases displayed an attention pattern similar
to true positives, with concentration in the same regions.
In true-negative cases, attention again concentrated on the QRS complex, with little attention over the adjacent ST-T segment.
Detailed interpretation of these patterns, including ICD code analysis of
misclassified cases, is provided in the Discussion.
Subgroup-level attention maps stratified by atrial fibrillation status, age
group, and sex are provided in Supplementary
Figures~\ref{fig:attn_afib}--\ref{fig:attn_sex}.

The averaged Lead~II signal reveals differences across prediction categories:
false-negative cases show a taller and narrower R peak, while the QRS complex
appears broader in true-positive and false-positive cases. A T-wave is more visible
following the QRS in false-negative cases, whereas it is less apparent in
true-positive and false-positive cases; the ST segment remains relatively flat
across all categories.
In true-negative cases, the averaged signal shows a tall, narrow R peak with a clearly visible T-wave, consistent with the preserved ventricular function expected in this group.

% ── 5.8 Decision Curve Analysis ──────────────────────────────────────────────

\subsection*{Decision Curve Analysis}

\begin{figure}[htbp]
    \centering
    \includegraphics[width=0.95\textwidth]{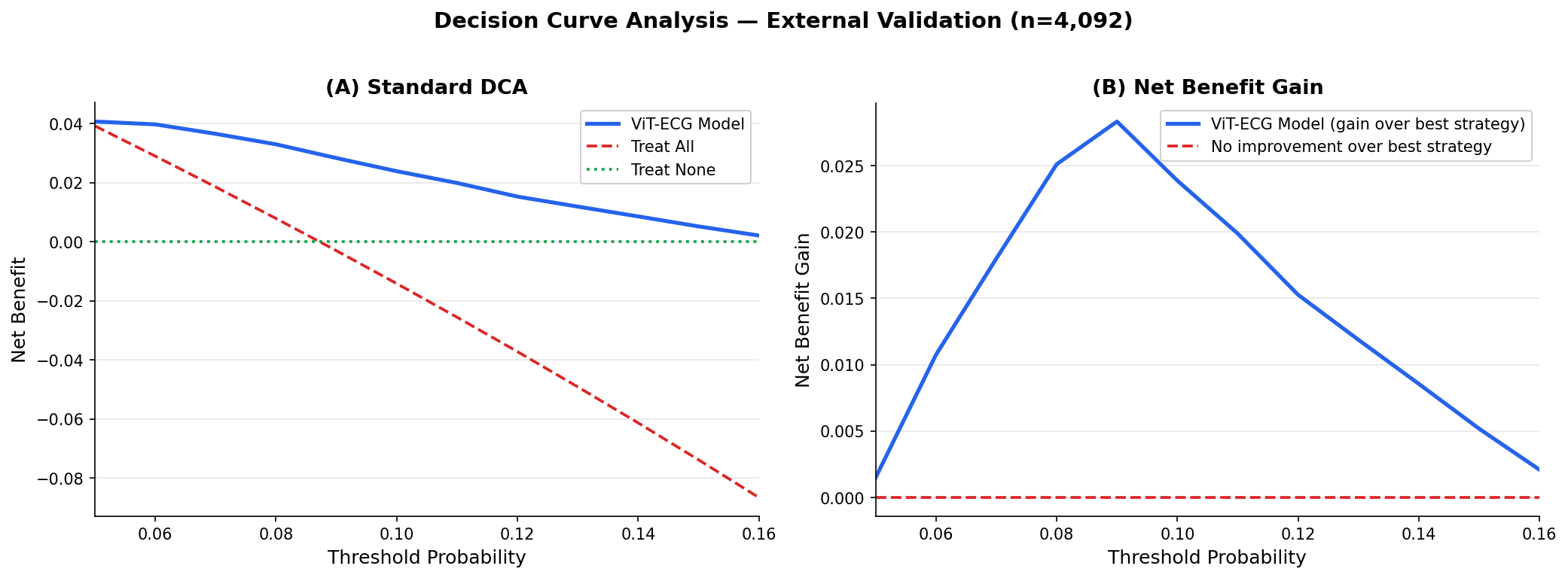}
    \caption{\textbf{Decision curve analysis for the ViT-ECG ensemble (external
    validation, n\,=\,4,092).}
    Panel (A) shows the standard DCA: net benefit across threshold probabilities
    from 5\% to 16\% for the ViT-ECG model (blue), treat-all (red dashed), and
    treat-none (green dotted).
    Panel (B) shows the net benefit gain: the improvement of the model over the
    best simple strategy at each threshold (treat-all where its net benefit is
    positive, treat-none otherwise).}
    \label{fig:dca}
\end{figure}

Decision curve analysis demonstrated that the ViT ensemble provided net clinical
benefit over both the treat-all and treat-none strategies across threshold
probabilities from 5\% to 16\% in the external validation set
(Figure~\ref{fig:dca}, Panel A). The net benefit gain peaked at approximately
8--9\%, corresponding to the observed prevalence of 8.72\%, where the model
achieves its greatest advantage over the best simple strategy
(Figure~\ref{fig:dca}, Panel B).

\subsection*{Ablation Study}

Training a ViT from scratch on 10,142 patients at a real-world
prevalence of 8.72\% presents two challenges: class imbalance and overfitting
on limited data. To address these, the training pipeline incorporated class
weighting and oversampling for the former, and label smoothing, L2
regularization, and data augmentation for the latter. The contribution of each
component was evaluated by training 3-model ensemble variants of the full ViT
with one component removed at a time. Each variant used identical architecture
and hyperparameters, differing only in the component under evaluation. Results
are reported on the external validation set at threshold = 0.45
(Table~\ref{tab:ablation}).
$\Delta$AUC represents the benefit of each component relative to the full
ensemble reference (positive = component improves performance).
Patch size had the largest effect on performance. Using a patch size of 25
instead of the selected 50 reduced AUROC by 1.43 points, while a patch size
of 100 reduced it by 0.37 points. Among training components, label smoothing
contributed the most ($+$1.35 AUROC points), followed by L2 regularization
($+$0.96), augmentation ($+$0.77), class weighting ($+$0.31), and oversampling
($+$0.25).

\begin{table}[htbp]
\centering
\caption{\textbf{Ablation study: contribution of each training component.}
$\Delta$AUC = AUC(full ensemble) $-$ AUC(variant), expressed in AUROC
percentage points, representing the benefit
of including each component (positive = component improves performance).
Each variant is a 3-model ensemble with one component removed.
Threshold = 0.45. External validation set.
CW = class weighting; LS = label smoothing; L2 = L2 regularization;
Aug = augmentation ratio; OSF = oversampling factor; PS = patch size.
A check mark indicates the component was included; a cross indicates it was removed.}
\label{tab:ablation}
\renewcommand{\arraystretch}{1.15}
\setlength{\tabcolsep}{4pt}
\small
\begin{tabular}{lccccccrrrl}
\toprule
\textbf{Variant} & \textbf{CW} & \textbf{LS} & \textbf{L2} & \textbf{Aug} & \textbf{OSF} & \textbf{PS} & \textbf{Sens} & \textbf{Spec} & \textbf{AUC} & \textbf{$\Delta$AUC} \\
\midrule
\textbf{Full ensemble (reference)} & \checkmark & \textbf{0.1} & \checkmark & \textbf{0.15} & \textbf{2} & \textbf{50} & \textbf{81.2\%} & \textbf{81.0\%} & \textbf{0.8775} & \textbf{—} \\
Label smoothing    & \checkmark & 0.0 & \checkmark & 0.15 & 2 & 50 & 76.8\% & 79.9\% & 0.8640 & $+$1.35 \\
L2 regularization  & \checkmark & 0.1 & $\times$   & 0.15 & 2 & 50 & 79.3\% & 78.9\% & 0.8679 & $+$0.96 \\
Augmentation       & \checkmark & 0.1 & \checkmark & 0.0  & 2 & 50 & 79.8\% & 79.7\% & 0.8698 & $+$0.77 \\
Class weight       & $\times$   & 0.1 & \checkmark & 0.15 & 2 & 50 & 73.7\% & 85.7\% & 0.8744 & $+$0.31 \\
Oversampling       & \checkmark & 0.1 & \checkmark & 0.15 & 1 & 50 & 75.6\% & 84.6\% & 0.8750 & $+$0.25 \\
Patch size = 25    & \checkmark & 0.1 & \checkmark & 0.15 & 2 & 25 & 77.9\% & 80.1\% & 0.8632 & $+$1.43 \\
Patch size = 100   & \checkmark & 0.1 & \checkmark & 0.15 & 2 & 100 & 81.2\% & 78.2\% & 0.8738 & $+$0.37 \\
\bottomrule
\end{tabular}
\end{table}

% ── 5.6 CNN Baseline Comparison ───────────────────────────────────────────────
\subsection*{CNN Baseline Comparison}

Four CNN configurations were compared against the ViT ensemble on the same
training and validation data (Table~\ref{tab:cnn_comparison}).
The vanilla CNN trained on full 5,000-step signals achieved an AUROC of
0.726 on the external validation set.
The best single-model CNN, trained on beat-level 1,500-step input, achieved an AUROC of 0.860, 78.4\% sensitivity,
and 77.3\% specificity.
When trained as a 3-model ensemble with ViT-matched parameters, sensitivity
increased to 95.8\% but specificity fell to 40.4\%.
The best CNN ensemble (no oversampling or augmentation)
achieved an AUROC of 0.853, 79.6\% sensitivity, and 77.6\% specificity.
The ViT ensemble achieved the highest AUROC of all configurations on both
validation sets (Table~\ref{tab:cnn_comparison}). It outperformed the best CNN
ensemble on the external
validation set (0.878 vs.\ 0.853; difference 0.025, 95\% CI
0.012--0.038; DeLong's test, $p = 1.5\times10^{-4}$) and on the internal
validation set (0.880 vs.\ 0.854; difference 0.026, 95\% CI 0.012--0.040; $p = 2.1\times10^{-4}$).

\begin{table}[htbp]
\centering
\caption{\textbf{CNN baseline configurations vs.\ ViT ensemble.}
All metrics at threshold\,=\,0.45.
OSF\,=\,oversampling factor;
Aug\,=\,augmentation ratio; LS\,=\,label smoothing.}
\label{tab:cnn_comparison}
\renewcommand{\arraystretch}{1.15}
\setlength{\tabcolsep}{4pt}
\small
\begin{tabular}{lccccccccc}
\toprule
\multirow{2}{*}{\textbf{Model}} &
\multirow{2}{*}{\textbf{OSF}} &
\multirow{2}{*}{\textbf{Aug}} &
\multirow{2}{*}{\textbf{LS}} &
\multicolumn{3}{c}{\textbf{External}} &
\multicolumn{3}{c}{\textbf{Internal}} \\
\cmidrule(lr){5-7}\cmidrule(lr){8-10}
& & & &
\textbf{Sens} & \textbf{Spec} & \textbf{AUC} &
\textbf{Sens} & \textbf{Spec} & \textbf{AUC} \\
\midrule
CNN vanilla (5{,}000 signals)       & 1 & 0    & 0   & 77.3 &  51.7 & 0.726 & 76.3 &  50.7 & 0.688 \\
CNN (1{,}500, best single model)    & 1 & 0    & 0   & 78.4 &  77.3 & 0.860 & 78.0 &  76.9 & 0.849 \\
CNN (1{,}500, ensemble, ViT params) & 2 & 0.15 & 0.1 & 95.8 &  40.4 & 0.867 & 96.5 &  40.7 & 0.868 \\
CNN (1{,}500, best ensemble)        & 1 & 0    & 0   & 79.6 &  77.6 & 0.853 & 77.4 &  77.7 & 0.854 \\
\midrule
\textbf{ViT (3-model ensemble)}     & 2 & 0.15 & 0.1 & \textbf{81.2} & \textbf{81.0} & \textbf{0.878} & \textbf{82.9} & \textbf{80.4} & \textbf{0.880} \\
\bottomrule
\end{tabular}
\end{table}

% Forest plot moved to supplementary
% \begin{figure}[htbp]
%     \centering
%     \includegraphics[width=\textwidth]{figures/subgroup_forest_plot.png}
%     \caption{\textbf{Subgroup performance forest plot — external validation.}
%     Sensitivity (blue) and specificity (red) with 95\% bootstrap confidence
%     intervals across demographic, EHR system, encounter setting, and
%     comorbidity subgroups.}
%     \label{fig:forest}
% \end{figure}

%% file: sections/discussion.tex
% ══════════════════════════════════════════════════════════════════════════════
% DISCUSSION
% ══════════════════════════════════════════════════════════════════════════════

\section*{Discussion}

Across 4,092 patients drawn from three geographically independent clinical sites,
a ViT trained from scratch on 10,142 patients achieved 81.2\%
sensitivity and 81.0\% specificity for detecting LVEF $\leq$40\%, with an AUROC
of 0.878. Although a PPV of 29.1\% may appear low, it is expected at the observed
prevalence of 8.72\%. A particularly relevant metric here is the NPV of 97.8\%. When the model clears
a patient, the probability of a missed reduced LVEF is about 2 in 100, supporting a rule-out function. Performance on the 3,295-patient internal
validation set was consistent (AUROC 0.880), reinforcing that the findings were
not specific to one set of sites.
In practice, such a model could serve as a first-pass triage layer applied
to routinely acquired ECGs, flagging patients for echocardiographic evaluation.

Performance was consistent across sex and care setting. Among racial and ethnic
groups with sufficient sample sizes, AUROC ranged from 0.849 in White,
non-Hispanic patients to 0.918 in Black or African American patients, with
Hispanic patients achieving 0.917. Asian, American Indian or Alaska Native,
Native Hawaiian or Other Pacific Islander, and Other or Multiple Race patients
each showed 100\% sensitivity. However, these groups had fewer than 10 positive
cases each, and their estimates should be interpreted with caution. These findings
are relevant to equitable deployment of AI-based clinical tools, where
performance disparities across demographic groups are a recognized concern.

A notable source of variation was specificity across age groups, which
declined from 89.3\% in patients under 60 to 72.3\% in those over 75.
Sensitivity, by contrast, was highest in the oldest group at 86.6\%. Age-related ECG changes
can produce morphological patterns that overlap with those associated with
reduced ejection fraction, making older patients an expected source of false
positives. In this subgroup, the elevated false-positive rate carries a practical consequence: unnecessary echocardiographic referrals impose real costs in resource use and patient burden. Specificity was also reduced in patients with atrial fibrillation
(70.2\%) and cardiomyopathy (57.0\%), both of which produce ECG changes that
can resemble reduced LVEF patterns.

Attention maps show which parts of the ECG the model focused on when predicting
reduced LVEF (Figure~\ref{fig:attention}). Across all four categories, two
findings emerge: the model attended to physiologically meaningful regions of the
ECG, and its misclassifications were not random but reflected the patients'
underlying cardiac characteristics. In true-positive cases, attention was
concentrated over the QRS complex, with weaker, secondary attention over the
adjacent ST-T segment.
In true-negative cases, attention remained centered on the QRS complex, indicating that the model relied on the same region to confirm preserved ventricular function as it did to flag reduced function.

False-negative cases showed attention concentrated on the QRS complex but little
attention over the ST-T segment and T wave, consistent with the model reading
these ECGs as structurally normal despite the underlying reduced LVEF. The model
therefore tended to miss patients who lacked its expected structural markers.
ICD code analysis of the 67 false-negative patients supports this: only 7\% had
a recorded conduction abnormality (ICD-10 I44), compared with 27\% of
true-positive patients, and only 70\% had a heart failure diagnosis versus 96\%
of true positives. These patients had confirmed reduced LVEF but were less
likely to carry the structural ECG markers that the model most reliably
associates with systolic dysfunction. In addition, 18 of the 67 false-negative
patients (27\%) had borderline EF in the 36--40\% range, where ECG changes are
inherently subtle.

False positives followed the opposite pattern: they were largely patients with
genuine structural disease whose ECGs resemble
those seen in reduced LVEF. Their attention concentration matched that of true positives.
ICD analysis of the 708 false-positive patients showed that
55\% had a heart failure diagnosis compared with 25\% of
true-negative patients, 45\% had atrial fibrillation versus 25\%, and 18\% had a
conduction abnormality versus 6\%. Continuous EF analysis reinforced this: mean
EF was 58.0\% and 76.1\% had EF above 50\%, indicating that false positives were
not primarily driven by borderline EF values near the 40\% threshold. Nearly
half of false-positive patients (49.7\%) had heart failure with preserved
ejection fraction, suggesting the model detects ECG patterns of structural
remodeling even when LVEF remains above the 40\% threshold. These maps are
population-level summaries averaged across patients and attention heads and
should not be used for individual clinical interpretation.

Subgroup-level attention maps (Supplementary Figures~\ref{fig:attn_afib}, \ref{fig:attn_age}, \ref{fig:attn_sex}) extend this analysis to clinically relevant patient groups.
In AFib patients, attention extended continuously across the QRS complex and the ST-T segment, whereas in non-AFib patients it concentrated more sharply on the QRS complex and T wave, with little attention to the intervening ST segment. The elevated false-positive rate in AFib patients is consistent with the frequent coexistence of AFib and structural heart disease, whose ECGs resemble those of reduced LVEF.
In the oldest age group, attention in false-positive cases was broad and diffuse, closely resembling the pattern in true positives. This is consistent with age-related ECG changes that the model interprets as reduced LVEF.
Attention structures were broadly similar between female and male patients, consistent with the comparable performance across sex.
Across all subgroups, the QRS complex remained the focus of attention while the P-wave region received consistently low attention, indicating the model relied on a stable, generalizable ventricular signal.

As a benchmark, we reimplemented the CNN architecture of
Attia et al.~\cite{attia2019}, later validated at scale by Carter
et al.~\cite{carter2026} and well established for ECG-based LVEF detection,
and trained it on the same data and splits as the ViT.
On the external validation set, the ViT ensemble outperformed the best CNN
ensemble by 2.5 AUROC points (0.878 vs.\ 0.853), with more balanced
sensitivity and specificity.
The gap narrowed to 1.0 point only when the CNN was tuned with the ViT's
hyperparameters, and even then the CNN matched the ViT only through a clinically
impractical operating point: a sensitivity of 95.8\% with a specificity of only
40.4\%. At this operating point, 6 in 10 patients without reduced LVEF would be
referred for unnecessary echocardiography.
Vision transformers are typically data-intensive, often requiring very large
training sets to perform well~\cite{dosovitskiy2021}. Despite training on a
relatively modest dataset of 10,142 patients, our ViT achieved strong
performance, suggesting that the approach is well suited to this task and may
improve further with larger training sets.

The ablation study shows that training decisions, not architecture alone, drove
the final performance. Patch size had the largest single effect: a size of 50,
corresponding to 100\,ms at 500\,Hz, best captured the temporal scale of
clinically relevant ECG features such as the QRS complex and T-wave. Smaller
patches fragmented these features and larger patches obscured the morphological
detail the model relies on. Among training components, the three largest
contributors were label smoothing ($+$1.35 AUROC points), L2 regularization
($+$0.96), and augmentation ($+$0.77). All three are forms of regularization,
suggesting that preventing overfitting was the central challenge when training
from scratch on 10,142 patients. These regularization components are what make
the ViT viable in a limited-data clinical
setting. The class weighting result illustrates a broader point: removing it
reduced AUROC by only 0.31 points, but sensitivity fell from 81.2\% to 73.7\%
while specificity rose to 85.7\%. In a triage context where missing cases
carry higher clinical cost than unnecessary referrals, AUROC alone can mask
clinically meaningful shifts in the operating point under class imbalance; we
therefore report sensitivity and specificity at a fixed threshold alongside
AUROC throughout.

Decision curve analysis showed that using the model to guide referral decisions
outperformed both referring all patients and referring no patients across
threshold probabilities of 5--16\% (Figure~\ref{fig:dca}). In this framework,
the threshold $p_t$ represents the minimum predicted probability at which a
clinician would refer a patient for echocardiography. It reflects a value
judgment about how harmful a missed diagnosis is compared with an unnecessary
echocardiogram. At $p_t = 0.10$, for instance, a clinician weighs a
missed LVEF $\leq$40\% diagnosis as nine times more harmful than an unnecessary
echocardiogram. This is a clinically reasonable position: echocardiography is
safe and non-invasive, while undetected heart failure carries risks of disease
progression and delayed treatment. By contrast, at $p_t = 0.90$, an unnecessary
echocardiogram is considered nine times more harmful than a missed diagnosis,
which is not clinically justifiable for LVEF detection.
Panel B of Figure~\ref{fig:dca} shows the net benefit gain, which measures how
much the model improves over the best simple strategy at each threshold. At very
low thresholds, the gain approaches zero because the model refers nearly all
patients, offering no advantage over referring everyone. The gain peaks at
approximately 8--9\%, corresponding to the observed prevalence of 8.72\%, where
the model is most selective and adds the most clinical value.

The training pipeline used minority oversampling and class weighting to
prioritize sensitivity. As a result, the raw ensemble overestimates absolute
risk. This is a shift in the probability scale rather than a deeper problem. The
calibration slope was already close to ideal (1.06), and discrimination was
unaffected. A single post-hoc Platt step, fit on held-out internal data and
applied to the external set, was enough to restore calibration (Brier 0.121 to
0.061; calibration-in-the-large $-2.01$ to 0.02), and it left the AUROC and the
operating-point sensitivity and specificity unchanged. A deployment that needs
absolute risk estimates rather than a binary triage flag can apply this step
directly.

These findings should be interpreted in the context of several limitations.
The validation was retrospective; prospective
evaluation in a real clinical workflow remains an important next step.
Each patient was represented by a single ECG selected by proximity to
echocardiography, which may not capture temporal variability in cardiac
function. Echocardiographic LVEF carries its own measurement variability,
and borderline cases near the 40\% threshold may be misclassified for
reasons unrelated to model performance. The model was trained on a binary
LVEF threshold; whether a continuous prediction target offers additional
clinical utility remains an open question. The elevated false-positive rate in patients over 75 represents a subgroup-specific limitation that warrants targeted model refinement. We did not fine-tune
foundation models such as HeartBEiT~\cite{heartbeit2023},
ECGFounder~\cite{ecgfounder2025}, or RhythmBERT~\cite{wang2026rhythmbert} on the
same training data, so a direct head-to-head comparison with pretraining-based
approaches on identical data has not been established. Waveform-level analysis to identify the specific ECG morphological features driving model predictions remains an important direction for improving interpretability.

In conclusion, a vision transformer trained from scratch on a modest, clinically
accessible dataset detected reduced LVEF from the 12-lead ECG, generalizing
across multiple health systems and patient subgroups and outperforming an
established CNN baseline. These findings support the potential of routine ECGs as a scalable, first-pass
triage step to identify patients who should undergo echocardiography for reduced
ejection fraction.

%% file: sections/methods.tex
% ══════════════════════════════════════════════════════════════════════════════
% METHODS
% ══════════════════════════════════════════════════════════════════════════════

\section*{Methods}

% ── 4.1 Study Design & Ethical Approval ──────────────────────────────────────

\subsection*{Study Design and Data Use}

This retrospective, observational cohort study analyzed data derived from
Dandelion Health's proprietary repository of de-identified clinical data
collected from its consortium of US health systems. Prior to extraction from
each health system, the data were de-identified using privacy-preserving
methodologies developed for each data type and approved by expert determination
under the HIPAA Privacy Rule.

% ── 4.2 Dataset ───────────────────────────────────────────────────────────────

\subsection*{Dataset}

\paragraph{Data sources.}
De-identified ECG and echocardiographic data were obtained from three US
health systems (10 acquisition sites in total): Sharp HealthCare (5 sites,
San Diego, CA), Sanford Health (3 sites, Sioux Falls, SD), and Texas Health
Resources (2 sites, Fort Worth, TX), accessed via the Dandelion Health platform.
Data spanned encounters from 2016 onward and covered inpatient, outpatient,
and observation settings.
ECG waveforms were recorded using GE Healthcare and Philips systems.

\paragraph{Patient selection.}
Patients aged 22 years or older who had at least one standard 12-lead digital
ECG paired with a transthoracic echocardiogram (TTE) containing a quantitative
numeric LVEF value were eligible for inclusion.
The ECG was required to precede the echocardiogram, and both studies had to
be performed within 30 calendar days of each other \cite{carter2026, adedinsewo2020}.
Only one ECG--TTE pair per patient was retained; for patients with multiple
eligible pairs, the earliest qualifying echocardiogram was selected and paired
with the closest preceding ECG.
Only ECGs with a duration of at least 10 seconds and a sampling rate of
500\,Hz were included.

\paragraph{Ground Truth Label.}
LVEF was measured using the biplane method of disks (modified Simpson's rule)
from two-dimensional apical views, consistent with standard clinical practice
guidelines \cite{lang2015}.
Echocardiograms were interpreted by board-certified cardiologists at each
participating site as part of routine clinical care. 
Patients with LVEF $\leq 40\%$ were classified as having left ventricular
systolic dysfunction (LVSD; positive class), consistent with the 2022
AHA/ACC/HFSA and 2021 ESC heart failure guidelines \cite{mcdonagh2023,
heidenreich2022}.
Patients with LVEF $> 40\%$ were classified as negative.

\paragraph{Dataset splits.}
The full dataset was partitioned at the patient level into training,
internal validation, and external validation sets with no patient overlap across
splits.
The training set comprised 10,142 patients drawn from all available
training sites.
Three subsamples were independently drawn from the training set (each 80\% of
training patients); one model was trained per subsample, yielding three models
that were subsequently combined into an ensemble.
The internal validation set comprised 3,295 patients from 7 sites and was
used exclusively for hyperparameter selection and threshold optimization.
The external validation set comprised 4,092 patients from 3 clinical sites,
each a physically and operationally independent location not used during training
or internal validation, with one site drawn from each of the three participating
health systems. These sites were entirely withheld from all model development
activities.
The algorithm was locked prior to any analysis of the external validation set.
The overall prevalence of LVEF $\leq 40\%$ in the external validation set was
8.72\%.

% ── 4.3 ECG Preprocessing ─────────────────────────────────────────────────────

\subsection*{ECG Preprocessing}

Raw ECG waveforms were stored in XML format, recorded at
500\,Hz for a minimum duration of 10 seconds. Each recording comprised 8
directly acquired leads (I, II, V1--V6).
The four augmented limb leads were derived using standard formulas
\cite{kligfield2007}: Lead~III $=$ Lead~II $-$ Lead~I;
aVR $= -($ Lead~I $+$ Lead~II $)/2$;
aVL $= ($ Lead~I $-$ Lead~III $)/2$;
aVF $= ($ Lead~II $+$ Lead~III $)/2$.

\paragraph{Signal filtering.}
Prior to R-peak detection, each lead signal was filtered using a 5th-order
Butterworth high-pass filter with a 0.5\,Hz cutoff to remove baseline wander,
followed by a notch filter at 50\,Hz to suppress powerline interference~\cite{makowski2021}.

\paragraph{R-peak detection.}
R-peaks were detected in Lead II using the NeuroKit2 library
\cite{makowski2021}.
If R-peak detection failed in Lead II (fewer than two valid peaks detected),
the algorithm fell back to Lead I.
ECGs with no detectable R-peaks in either lead were excluded.

\paragraph{Normalization.}
Each lead was z-score normalized (zero mean, unit variance) over the full
recording, prior to beat-window extraction, to account for inter-patient
amplitude variability.

\paragraph{Beat-level RR segmentation.}
Non-overlapping beat windows were extracted by iterating through detected
R-peaks with a step size of 2: each window spanned from $R_i$ to $R_{i+2}$,
capturing two consecutive RR intervals \cite{mohan2024}.
Each window was truncated to a maximum of 1,500 time steps and zero-padded
at the beginning if shorter.
This frames the model's analysis at the level of individual heartbeats rather than the full ECG recording.

% ── 4.4 Model Architecture — Vision Transformer ───────────────────────────────

\subsection*{Model Architecture}
\begin{figure}[htbp]
    \centering
    \includegraphics[width=0.85\textwidth]{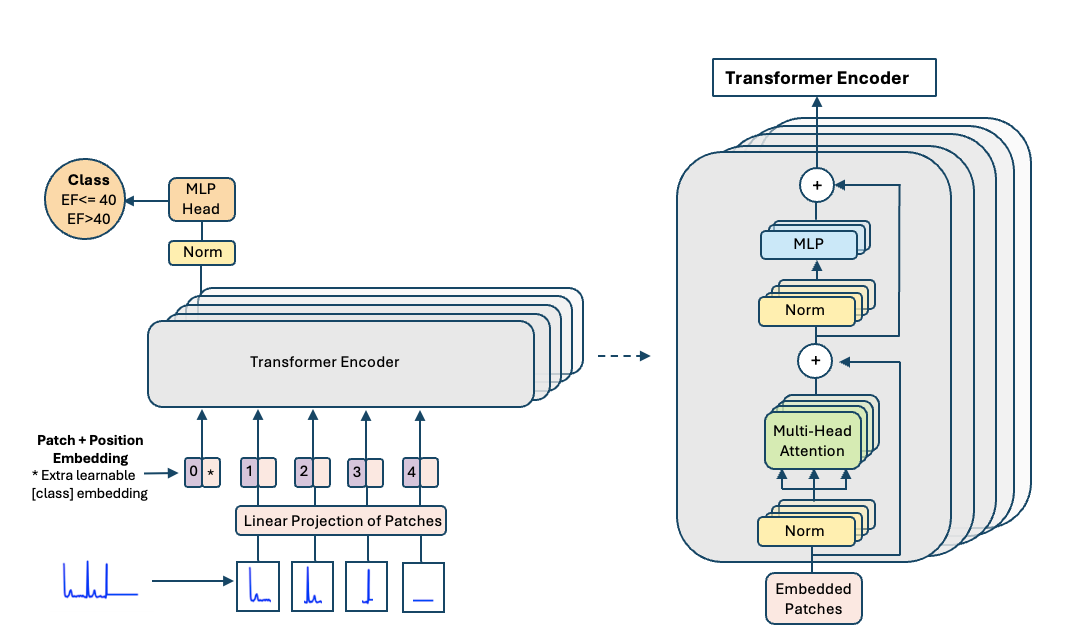}
    \caption{\textbf{Vision Transformer architecture for ECG-based LVEF classification.}
    ECG beat windows are divided into patches and projected to embeddings.
    A learnable CLS token is prepended, and the sequence is processed by Transformer Encoder
    layers with pre-norm multi-head self-attention and residual connections.
    The CLS token representation is passed to an MLP head to predict LVEF $\leq 40\%$ or $> 40\%$.}
    \label{fig:vit_arch}
\end{figure}

We adapted the Vision Transformer (ViT) architecture \cite{dosovitskiy2021}
for one-dimensional multi-lead ECG classification (Figure~\ref{fig:vit_arch}).
The model receives a single beat window as input: a matrix of shape
$1500 \times 12$ (time steps $\times$ leads).

\paragraph{Patch embedding.}
The 12-lead signal (shape $1500 \times 12$) was treated as a single
multi-channel sequence and divided into non-overlapping patches of size 50
along the time axis using a 1D convolutional projection, yielding 30 patches
in total ($1500 / 50 = 30$).
Each patch was projected to a 64-dimensional embedding vector.
A learnable class (CLS) token was prepended to the patch sequence, resulting
in 31 tokens per input.
Learnable positional embeddings of the same dimension were added to each
token to preserve temporal order.

\paragraph{Transformer encoder.}
The token sequence was processed by a stack of 3 transformer encoder layers,
each comprising multi-head self-attention (4 heads, key dimension $= 16$)
followed by a position-wise feed-forward network with GELU activations.
Layer normalization was applied before each sub-layer (pre-norm formulation).
L2 regularization ($\lambda = 10^{-4}$) was applied to the feed-forward and
classifier weights.

\paragraph{Classification head.}
After the final encoder layer, a layer normalization and an MLP with 64 units
(GELU activations, L2 regularization, dropout $= 0.3$) were applied to the
full token sequence.
The representation of the CLS token was then extracted and passed to a
2-unit Dense layer with softmax activation, producing class probabilities
for LVEF $\leq 40\%$ (positive) and LVEF $> 40\%$ (negative).

% % ── 4.5 Training Strategy ─────────────────────────────────────────────────────

\subsection*{Training Strategy}
\label{sec:training}

\paragraph{Loss function and class imbalance.}
Models were trained with categorical cross-entropy loss.
To address the imbalance between negative (LVEF $> 40\%$, 77.5\%) and
positive (LVEF $\leq 40\%$, 22.5\%) classes in the training set, two
complementary strategies were applied: (1) class weighting, with the positive
class assigned a weight of 1.52 relative to the negative class; and (2)
oversampling of the minority class by a factor of 2, introducing additional
copies of positive-class beat windows at each epoch.

\paragraph{Data augmentation.}
To improve generalization, 15\% of training beats were randomly selected and
augmented by adding Gaussian noise (standard deviation 0.01) and applying
amplitude scaling (factor uniformly sampled from $[0.9, 1.1]$).
The augmented beats were added to the training set as additional copies, and the
original beats were retained.

\paragraph{Optimizer and regularization.}
Models were optimized using Adam (learning rate $= 10^{-4}$,
gradient clip value $= 0.5$).
The learning rate was reduced by a factor of 0.3 when the validation loss
plateaued for 5 consecutive epochs (ReduceLROnPlateau, minimum
$\mathrm{lr} = 10^{-6}$).
Training was stopped early when the validation loss did not improve for 8
consecutive epochs, restoring the weights from the best epoch (maximum 50
epochs, batch size 32).

\begin{figure}[htbp]
    \centering
    \includegraphics[width=0.55\textwidth]{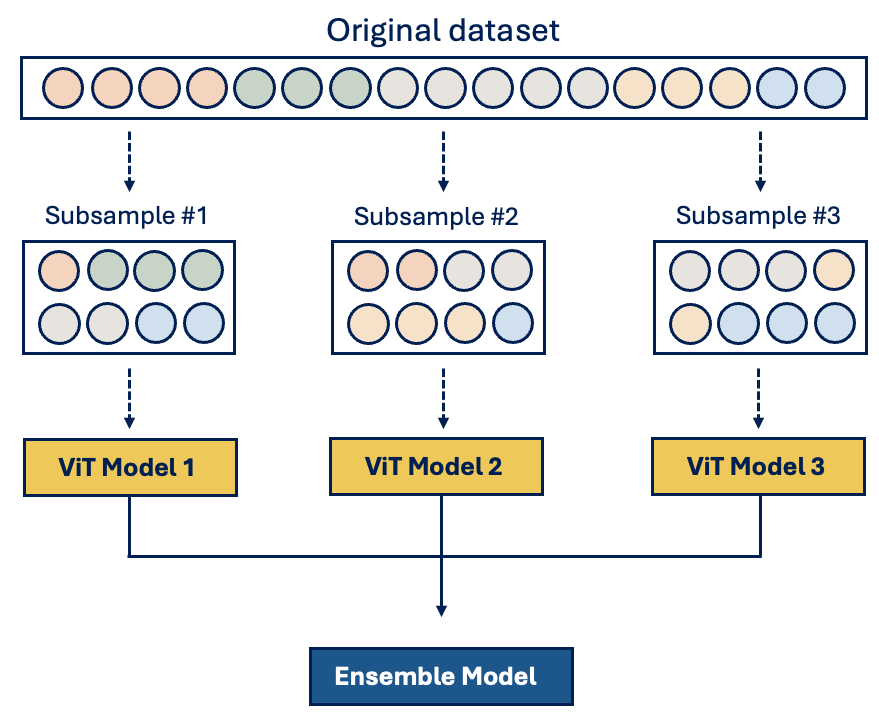}
    \caption{\textbf{Ensemble training strategy.}
    Three ViT models were trained on independent random subsamples of the training data
    and combined into an ensemble by averaging their predicted probabilities.}
    \label{fig:ensemble}
\end{figure}

\paragraph{Ensemble.}
Three subsamples were independently drawn from the training set at the patient
level (each comprising 80\% of available training patients, stratified by label).
One ViT model was trained independently on each subsample, with early stopping
guided by the internal validation set.
At inference, the three models were combined into an ensemble by averaging
their predicted positive-class probabilities (Figure~\ref{fig:ensemble}).
The ensemble probability was then thresholded to produce a binary prediction
(threshold selection described in the Threshold Selection section).

% % ── 4.6 CNN Baseline ──────────────────────────────────────────────────────────

\subsection*{CNN Baseline}

To benchmark ViT performance against a well-established architecture, we
reimplemented the convolutional neural network (CNN) described by Attia
et al.\ \cite{attia2019} as a comparison baseline.
The architecture consists of six temporal convolutional blocks applied across
the time dimension, followed by one spatial convolutional block applied across
the lead dimension, and two fully connected layers.
Blocks 1--3 use a kernel size of 5; blocks 4--6 use a kernel size of 3.
Filter counts increase from 16 in blocks 1--2 to 32 in blocks 3--4 and 64 in
blocks 5--6.
Max-pooling with stride 4 is applied after blocks 3 and 5, and max-pooling
with stride 2 is applied after all other temporal blocks.
The spatial block uses a pointwise convolution (kernel size 1) to combine
information across leads.
The two fully connected layers have 64 and 32 units respectively, each followed
by ReLU activation, batch normalization, and dropout.

Four CNN configurations were evaluated (Table~\ref{tab:cnn_comparison}).
The first used the full 12-lead waveform (5,000 time steps per patient,
matching the format of Attia et al.~\cite{attia2019}) with no class weighting,
oversampling, augmentation, or label smoothing, trained as a single model.
The remaining three configurations used the same RR-segmented 1,500-step
beat-level input as the ViT, with patient-level predictions obtained by
averaging per-beat probabilities.
Among these: the best single-model configuration used a positive class weight
of 2.3 with no oversampling, augmentation, or label smoothing; the best
ensemble configuration used class weight 1.16 across three models; and a
further ensemble configuration matched the core ViT training parameters
(class weight 1.52, oversampling factor 2, augmentation ratio 0.15,
label smoothing 0.1).
All RR-segmented configurations used a classification threshold of 0.45.
Training parameters for the RR-segmented configurations (class weight,
oversampling factor, and augmentation ratio) were selected to achieve the best
balance between sensitivity and specificity on the internal validation set;
the configuration matching ViT training parameters used identical values to
those of the ViT ensemble.

% ── 4.7 Threshold Selection ───────────────────────────────────────────────────

\subsection*{Threshold Selection}
\label{sec:threshold}

The operating threshold for the ensemble positive-class probability was
selected on the internal validation set by maximizing the Youden index
(sensitivity + specificity $-$ 1). A threshold of 0.45 was identified and
locked prior to any analysis of the external validation set.

% ── 4.8 Statistical Analysis ──────────────────────────────────────────────────

\subsection*{Statistical Analysis}

Model performance was characterized using sensitivity, specificity, area under
the receiver operating characteristic curve (AUROC), accuracy, positive
predictive value (PPV), negative predictive value (NPV), and F1 score.
All metrics were computed at the patient level using the ensemble
probability at the pre-specified threshold of 0.45.
Ninety-five percent confidence intervals (CIs) were estimated by
bias-corrected and accelerated (BCa) bootstrap resampling (10,000 iterations,
patient-level sampling with replacement).
The AUROCs of the ViT ensemble and the CNN baseline were compared using
DeLong's test for two correlated ROC curves, with a two-sided $p < 0.05$
considered statistically significant.

Model calibration was assessed on both validation sets using the Brier score,
calibration curves (10 quantile bins), the calibration slope, and
calibration-in-the-large. Because the training pipeline used minority
oversampling and class weighting to prioritize sensitivity, the raw ensemble
probabilities were expected to overestimate absolute risk. We therefore also
applied post-hoc Platt scaling: a logistic recalibration model was fit on the
internal validation set, mapping the logit of the ensemble probability to the
observed label, and then applied to the external validation set. Platt scaling
is monotonic, so it leaves the AUROC and the binary classification at any
threshold unchanged and adjusts only the probability scale.

Decision curve analysis (DCA) was used to evaluate the net clinical benefit of
the model compared with treat-all and treat-none strategies across a range of
threshold probabilities \cite{vickers2006}. Net benefit was calculated as:
\begin{equation}
    \text{Net Benefit} = \frac{\text{TP}}{N} - \frac{\text{FP}}{N} \cdot \frac{p_t}{1 - p_t}
\end{equation}
where $N$ is the total number of patients, TP and FP denote true and false
positives at threshold $p_t$, and $p_t / (1 - p_t)$ is the harm weight assigned
to false positives relative to false negatives. The analysis was performed across
threshold probabilities of 5--16\%. Net benefit gain was additionally computed
as the improvement of the model over the best simple strategy at each threshold.

% ── 4.9 Attention Map Analysis ────────────────────────────────────────────────

\subsection*{Attention Map Analysis}

To explore the spatial focus of the model, attention maps were extracted from
the final transformer encoder layer of the 12-lead ViT.
For each beat, the CLS token attention scores over the patch sequence were
obtained from the last self-attention layer and averaged across the 4 attention
heads.
Attention weights were then resized to the original signal length, cropped to
remove pre-padding, and interpolated to 1,500 time steps.
Beat-level attention maps were averaged across all beats per patient and
subsequently averaged across all patients within each prediction category
(true positive, true negative, false negative, false positive).
This procedure was repeated across the 3 ensemble models and the resulting
attention profiles were averaged.
The displayed signal corresponds to Lead~II; attention weights were derived
from the full 12-lead input.
Attention maps are presented as an exploratory, post-hoc analysis and do not
constitute mechanistic causal attribution.

% ── 4.10 Subgroup Analysis ────────────────────────────────────────────────────

\subsection*{Subgroup Analysis}

Subgroup analyses were performed on the external validation set to assess
the consistency of model performance across clinically relevant partitions.
Subgroups included: sex (female vs.\ male); age group ($<$60, 60--75,
and $>$75 years); race and ethnicity; encounter setting (inpatient
vs.\ outpatient); acquisition site; and the presence or absence of the
following comorbidities: atrial fibrillation, hypertension, type~2
diabetes, coronary artery disease, acute myocardial infarction,
cardiomyopathy, heart failure, and obesity.
Sensitivity, specificity, and AUROC were reported for each subgroup.
Results were summarized in a table.

%% file: sections/appendix.tex
\section*{Supplementary Material}

\setcounter{figure}{0}
\renewcommand{\thefigure}{S\arabic{figure}}

\subsection*{Subgroup Attention Maps}

To further investigate model behavior across clinically relevant patient subgroups, we generated mean attention maps separately for each subgroup using the external validation set. For each subgroup, the attention signal was averaged across all patients within that subgroup and prediction category (true positive, false negative, false positive). A higher attention weight (darker red) indicates that the model focused more strongly on that region of the normalized RR segment when making its prediction.

\begin{figure}[h!]
    \centering
    \includegraphics[width=\textwidth]{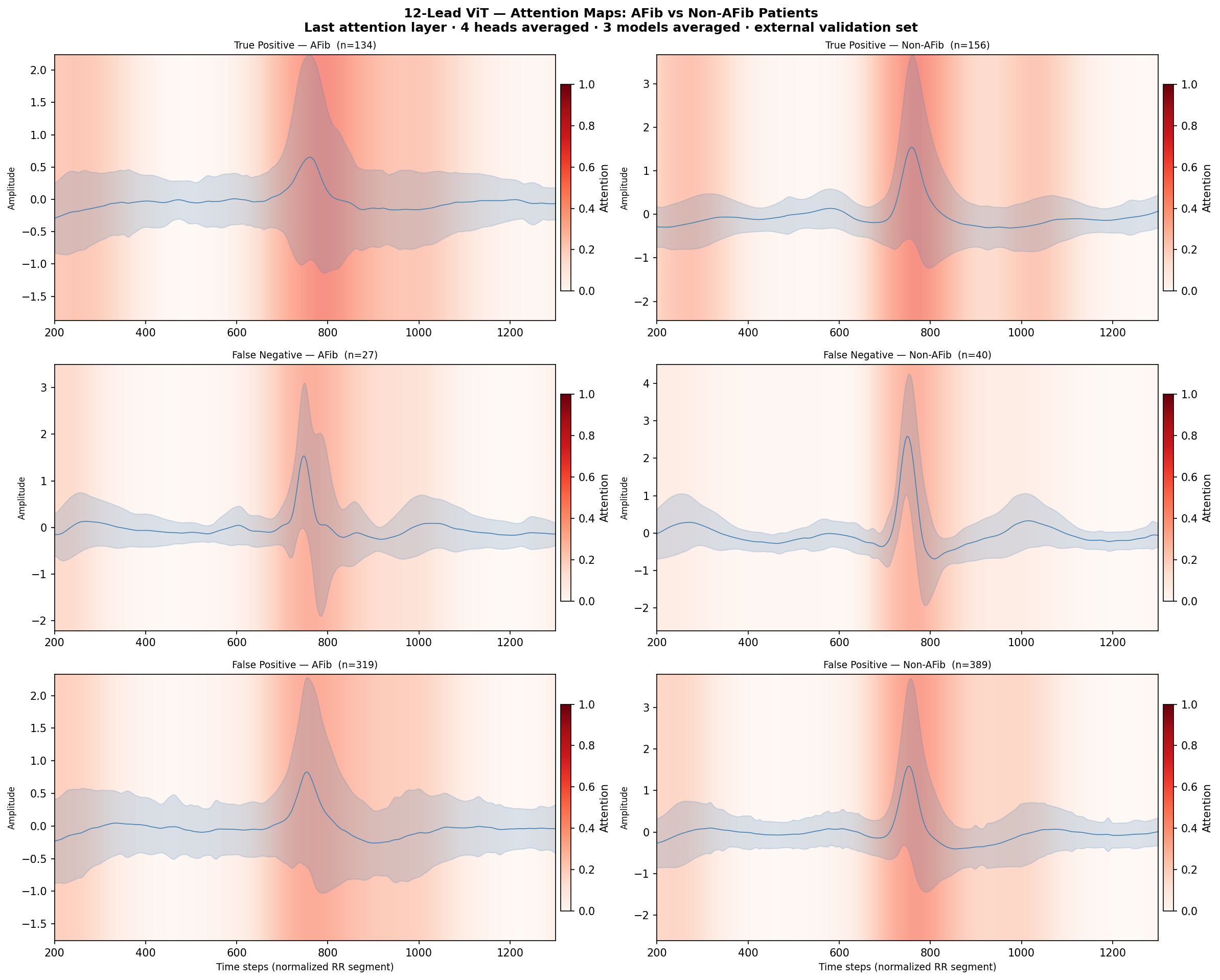}
    \caption{\textbf{Mean attention maps stratified by atrial fibrillation status.}
    Each row corresponds to a prediction category (true positive, false negative, false positive) and each column to AFib vs.\ non-AFib patients in the external validation set.
    The blue line and shaded band show the mean $\pm$ SD of the normalized RR-segmented ECG signal (Lead~II); the red heatmap overlaid on each panel shows the normalized mean attention weight from the last transformer layer (4 heads, 3 models averaged).
    The number of patients ($n$) contributing to each panel is shown in the subplot title.}
    \label{fig:attn_afib}
\end{figure}

\begin{figure}[h!]
    \centering
    \includegraphics[width=\textwidth]{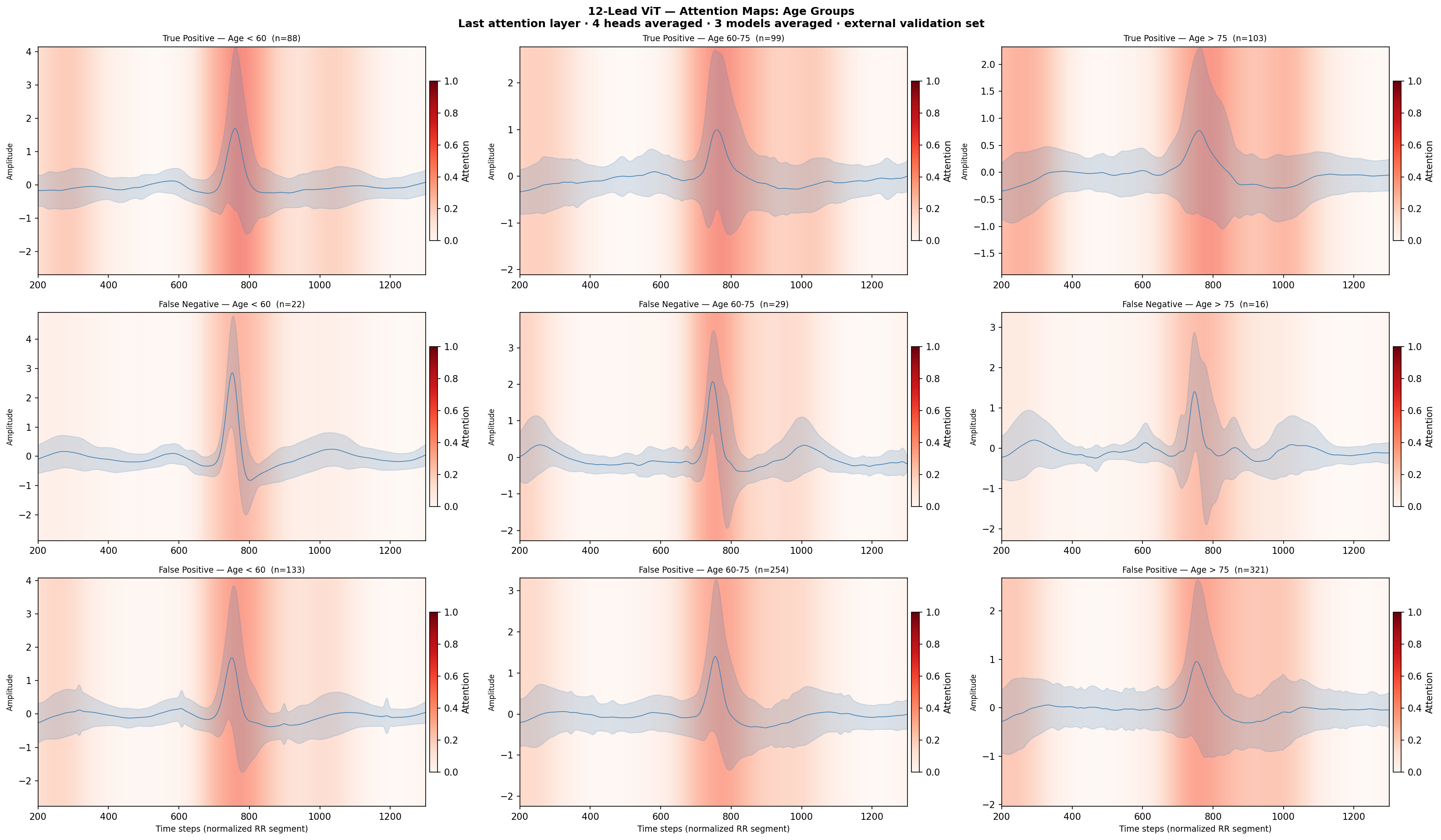}
    \caption{\textbf{Mean attention maps stratified by age group (${<}60$, $60$--$75$, ${>}75$ years).}
    Each row corresponds to a prediction category; columns correspond to the three age groups.
    Layout otherwise identical to Figure~\ref{fig:attn_afib}.}
    \label{fig:attn_age}
\end{figure}

\begin{figure}[h!]
    \centering
    \includegraphics[width=\textwidth]{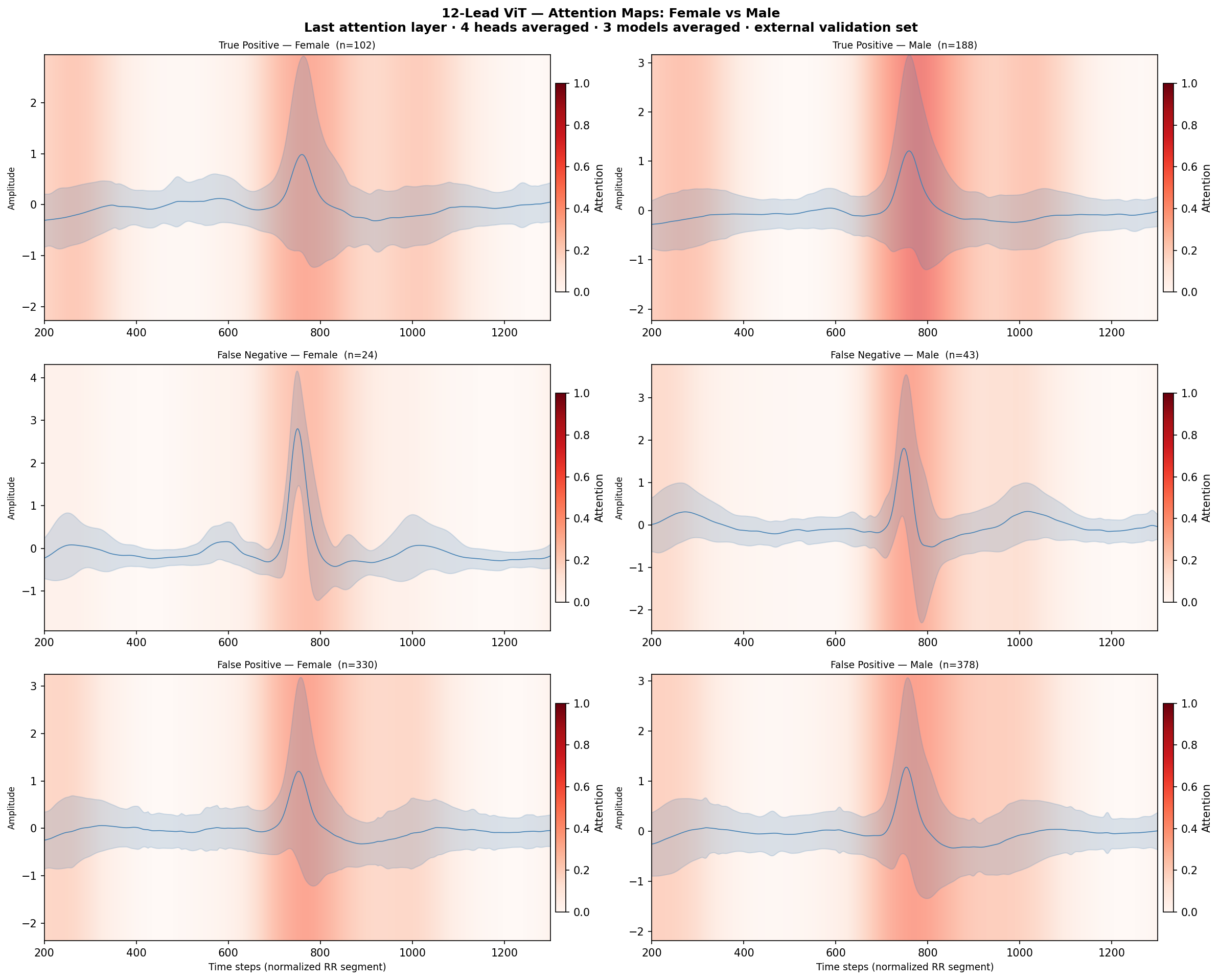}
    \caption{\textbf{Mean attention maps stratified by sex (female vs.\ male).}
    Layout identical to Figure~\ref{fig:attn_afib}.}
    \label{fig:attn_sex}
\end{figure}